\documentclass[autodetect-engine,dvipdfmx,10pt,a4paper,twocolumn]{article}

\usepackage{bm}
\usepackage{amsmath}
\usepackage{amssymb}
\usepackage{amsthm}
\usepackage[linesnumbered,ruled]{algorithm2e}
\usepackage{booktabs}
\usepackage{url}
\usepackage{framed}
\usepackage{multirow}

\usepackage{color}
\usepackage{colortbl}
\usepackage{breqn,flexisym,array}
\usepackage{dsfont}
\usepackage{graphicx}

\newtheorem{theorem}{Theorem}[section]
\newtheorem{lemma}{Lemma}

{\endMakeFramed}

\newenvironment{theorem-waku}
  {\begin{theorem}}
  {\end{theorem}}

\newenvironment{lemma-waku}
  {\begin{lemma}}
  {\end{lemma}}

\newcommand{\0}{{\bm{0}}}
\newcommand{\1}{{\bm{1}}}

\newcommand{\vf}{{\bm{f}}}

\newcommand{\vq}{{\bm{q}}}

\newcommand{\vu}{{\bm{u}}}

\newcommand{\vv}{{\bm{v}}}
\newcommand{\vw}{{\bm{w}}}

\newcommand{\x}{{\bm{x}}}

\newcommand{\y}{{\bm{y}}}

\newcommand{\z}{{\bm{z}}}

\newcommand{\vA}{{\bm{A}}}

\newcommand{\sfA}{{\textsf{A}}}

\newcommand{\vB}{{\bm{B}}}
\newcommand{\cB}{{\mathcal{B}}}
\newcommand{\sfB}{{\textsf{B}}}

\newcommand{\sfC}{{\textsf{C}}}

\newcommand{\cD}{{\mathcal{D}}}
\newcommand{\sfD}{{\textsf{D}}}

\newcommand{\cH}{{\mathcal{H}}}

\newcommand{\cI}{{\mathcal{I}}}

\newcommand{\vK}{{\bm{K}}}

\newcommand{\cM}{{\mathcal{M}}}

\newcommand{\bN}{{\mathbb{N}}}

\newcommand{\bR}{{\mathbb{R}}}

\newcommand{\cS}{{\mathcal{S}}}

\newcommand{\X}{{\bm{X}}}

\newcommand{\valph}{{\bm{\alpha}}}
\newcommand{\vbeta}{{\bm{\beta}}}

\newcommand{\vsig}{{\bm{\sigma}}}

\newcommand{\argmax}{\mathop{\textrm{argmax}}\limits}
\newcommand{\argmin}{\mathop{\textrm{argmin}}\limits}

\newenvironment{tsaligned}{\begin{equation}\begin{aligned}}{\end{aligned}\end{equation}}

\newenvironment{tsaligned*}{\begin{equation*}\begin{aligned}}{\end{aligned}\end{equation*}}

\newcommand{\tslong}[1]{{#1}}
\newcommand{\tsshort}[1]{{}}
\newcommand{\REFFIGDEMOCORA}{3}
\newcommand{\REFFIGDEMOMNIST}{4}
\newcommand{\REFFIGSVMPAIRWISE}{5}
\begin{document}
\title{Learning Sign-Constrained Support Vector Machines}
\author{%
Kenya Tajima, Takahiko Henmi, Kohei Tsuchida, \\
Esmeraldo Ronnie R.~Zara, and Tsuyoshi Kato
\footnote{%
Faculty of Science and Technology, Gunma University, 
Tenjin-cho 1-5-1, Kiryu, Gunma 376-8515, Japan. 
}
}

\maketitle


\textbf{Abstract:} 
Domain knowledge is useful to improve the generalization performance of learning machines. Sign constraints are a handy representation to combine domain knowledge with learning machine. In this paper, we consider constraining the signs of the weight coefficients in learning the linear support vector machine, and develop two optimization algorithms for minimizing the empirical risk under the sign constraints. One of the two algorithms is based on the projected gradient method, in which each iteration of the projected gradient method takes $O(nd)$ computational cost and the sublinear convergence of the objective error is guaranteed. The second algorithm is based on the Frank-Wolfe method that also converges sublinearly and possesses a clear termination criterion. We show that each iteration of the Frank-Wolfe also requires $O(nd)$ cost. Furthermore, we derive the explicit expression for the minimal iteration number to ensure an $\epsilon$-accurate solution by analyzing the curvature of the objective function.  Finally, we empirically demonstrate that the sign constraints are a promising technique when similarities to the training examples compose the feature vector. 

\section{Introduction}
Support vector machine (SVM)~\cite{cristianini2000} is a popular methodology applicable to extensive domains requiring prediction or statistical analysis including text categorization, medical and biological study, economics, psychology and social science. The methodology learns the coefficient vector $\vw:=\left[w_{1},\dots,w_{d}\right]^\top$ of a linear classifier $\left<\vw,\cdot\right>$ from a set of examples, each of which consists in $d$ features and a binary class label. 

Usually, domain knowledge is exploited when designing features so that the selected features are possibly correlated to the binary category. In many cases, for each selected feature, the sign of the correlation to the binary category is known in advance. For example, when predicting the pathogen concentration in a river, we may use water quality indices. In the field of the water quality engineering~\cite{KatKobOis-jwh19,Varela18a,KatKobIto15a}, no one doubts the fact that EC, SS, BOD, TN, TP and WT are positively correlated to the concentration of \textit{Escherichia coli} (\textit{E. coli}) in a river, and DO and FW have a negative correlation~\cite{KatKobOis-jwh19}. The hydrogen exponent is also informative for \textit{E. coli} count prediction because the organism cannot survive in acidic or basic water. 

It would be ideal if the signs of coefficients $\vw$ coincided with the signs of true correlations to the binary class labels. In case that $h$-th feature variable $x_{h}$ is positively correlated to the class label, then a positive weight coefficient $w_{h}$ is expected to achieve a better generalization performance than a negative coefficient. However, the standard SVM learning cannot prevent the learned weight from being negative if the correlation is weak, and vice versa.

In this paper, we discuss constraining the signs of the coefficients explicitly in SVM learning.
A simple approach to the constrained optimization is the \emph{projected gradient} method~\cite{Bertsekas99} in which a step for projection onto the feasible region is inserted at each iteration of the gradient method. We focus on \emph{Pegasos} method~\cite{Shalev-Shwartz11a-pegasos}, a popular gradient method for SVM learning with the \emph{sublinear convergence} guaranteed.
We consider inserting the projection step into each of Pegasos iteration, and analyzed the number of iterations to ensure a solution $\vw$ to be $\epsilon$-\emph{accurate} (this term is defined in Section III). We have found that the required number of iterations to ensure $\epsilon$-accuracy is bounded by twice the iteration numbers taken by the original Pegasos method. 
The time complexity for each iteration of the Pegasos-based method is $O(nd)$, which is equal to that for the original Pegasos.
The projected gradient method suffers from a drawback which is the incapability of the judgement for attaining $\epsilon$-accurate solution.

We present an alternative algorithm that possesses the clear termination criterion guaranteeing the $\epsilon$-accuracy for the solution. 
The alternative learning algorithm does not solve the primal problem directly, but solves the dual problem of the sign constrained SVM learning problem. To maximize the dual objective function, the \emph{Frank-Wolfe} framework~\cite{Levitin66,jaggi13icml} is adopted, which allows the proposed algorithm to inherit the sublinear convergence property from the Frank-Wolfe framework. 
Each iteration of the Frank-Wolfe framework consists of two steps: the \emph{direction finding} step and the \emph{line search} step. In this study, we have obtained the following findings: In the Frank-Wolfe algorithm for solving the dual problem to SVM learning under sign constraints, 
both the direction finding and line search steps can be performed within $O(nd)$ computational cost.
Hence, we have reached an analytical result that each iteration of the Frank-Wolfe algorithm takes $O(nd)$ computational cost which is same as the time complexity of each iteration in the projected gradient algorithm. This suggests that the proposed Frank-Wolfe algorithm totally shares all the advantages of the projected gradient algorithm, and furthermore overcomes a drawback: lack of the termination criterion. 

Exploitation of domain knowledge is a straightforward application of sign constraints, although sign constraints is effective when utilizing exploitation of a task structure.

We found that the sign constraints are advantageous to the task structure of the biological sequence classification~\cite{LiaNob03-jcb,KatTsuAsa05a}. 
Predicting some properties such as structural similarity and cellular functionality of a protein is a central issue in bioinformatics to understand the mechanism in the cell~\cite{KinKatTsu04a,KatNag10a}. A protein is a sequence of amino acids. More similar amino-acid sequences share a common ancestor with higher probability, tending to have common properties such as structural similarity and cellular function. Hence, sequence alignment is a typical methodology to infer the properties of unknown proteins. In this study, we found that combining sequence similarities with sign-constrained SVM provides an effective methodology for protein function inference. We shall demonstrate the efficiency on the new application of sign-constrained SVM to protein function inference.

This paper is organized as follows: After discussing related work in next section, the learning problem to be minimized is formulated in Section~\ref{s:scsvm}. The projected gradient algorithm and the analysis are analyzed in Section~\ref{s:pega}. In Section~\ref{s:dual}, the dual problem is presented and, in Section~\ref{s:stdfw}, the Frank-Wolfe algorithm for solving the dual problem is analyzed. Experimental results for convergence behaviors and the application to biological sequence classification are presented in Section~\ref{s:exp}. In the last section, this paper is concluded with future work for exploring the potential of sign constraints. 
\section{Related Work}
\label{s:related}
For discriminative learning, use of sign constraints has not been discussed well so far, although non-negative least square regression for other applications has been explored extensively.
The applications of non-negative least square regression include non-negative image restoration~\cite{Henrot2013-icassp}, 
face representation~\cite{YangfengJi2009-icmla,HeZheHu13}, microbial analysis~\cite{CaiGuKen17}, image super-resolution~\cite{DonFuShi16}, spectral analysis~\cite{QiangZhang07-asrc}, tomographic imaging~\cite{JunMa2013-algo}, and sound source localization~\cite{YuanqingLin2004-icassp}. 
Some of them used non-negative least square regression as a key ingredient for the non-negative matrix factorization~\cite{lee2001algorithms}. 
Non-negative least squares estimation can be computed efficiently. A basic algorithm for the non-negative least regression is the active set method~\cite{Lawson1995solving} which is accelerated by integrating projected gradient approach~\cite{DongminKim2010-siamjsc}. 
These algorithms are computationally stable and fast. However, these approaches help to minimize the empirical risk only for the case that the loss function is the square error. 

We are aware of an existing report in which prior knowledge for coefficients of linear classifiers is exploited~\cite{Fernandes2017}, as with our approach. They add a term penalizing the violation to the prior knowledge. The penalizing strength is controlled by a constant coefficient.
The learning problem discussed in our study is an extreme case of their formulation~\cite{Fernandes2017}. 
They employ a gradient method for minimizing the penalized empirical risk, although the details of the algorithm were not provided.
This paper focuses on how to minimize the empirical risk under sign constraints in well-appointed framework, to rigorously analyze the convergence rate and the computational cost. 
Without sign constraints, many optimization algorithms are available for generic empirical risk minimization~\cite{Roux12a-sag,Schmidt2016-sag,Johnson13a-svrg,LinXiao2014-siamjo,Defazio2014-nips}.
Many of them require a step size which is in need to be chosen manually in advance, although theoretically guaranteed step sizes are too small to converge the solution to the minimum for practical use.
The Frank-Wolfe algorithm developed in this paper works without any step-size as well as any hyper-parameter for optimization. 

Proposal for use of sign constraints in biological sequence analysis is another contribution of this paper. To find the evidence for homology between two biological sequences, sequence similarity has been used frequently~\cite{SmiWat81,AltGisMil90,Pea90}. Liao et al~\cite{LiaNob03-jcb} devised a method, named \emph{SVM-pairwise}, that combines the sequence comparison approach with SVM. In SVM-pairwise, the sequence similarities to all known sequences are used to train SVM as a feature vector. Supported with the remarkable success of SVM-pairwise, a large number of similar techniques have been developed for different tasks in the computational biology~\cite{LanDenCri04,LanBieCri04,LiuZhaXu14,OguMum06,KatTsuAsa05a}. In this paper, it is demonstrated that the generalization power of SVM-pairwise can be improved significantly by exploiting the effect of sign constraints.

\section{Sign-Constrained SVM}
\label{s:scsvm}
In this section, we consider imposing sign constraints for SVM learning. The sign constraints are posed for the coefficients for which the signs of the correlations between the corresponding features and the class labels are available in advance. Let us denote by $\cI_{+}$ and $\cI_{-}$ the index set of features known to be correlated to the class label positively and negatively in advance, where $\cI_{+}\cap\cI_{-}=\emptyset$ and $\cI_{+}\cup\cI_{-}\subseteq [d]$. The sign constraints are defined as
\begin{tsaligned}
  & \forall h\in\cI_{+}, \,\, w_{h}\ge 0
  \quad
  \text{and}
  \quad
  \forall h'\in\cI_{-}, \,\, w_{h'}\le 0. 
\end{tsaligned}
We may eliminate the non-positive constraints because these constraints can be transformed to the non-negative constraints by negating features for $\cI_{+}$ in advance. Hereinafter, we assume that this preprocess is performed in advance. Then, $\cI_{-}=\emptyset$. We introduce a constant vector $\vsig:=\left[\sigma_{1},\dots,\sigma_{d}\right]^\top\in\{0,1\}^{d}$ such that
\begin{tsaligned}
  \sigma_{h}:=\mathds{1}[h\in\cI_{+}]
\end{tsaligned}
where $\mathds{1}[x]=1$ if the logical argument $x$ is true; otherwise, $\mathds{1}[x]=0$. Then, the sign constraints can be rewritten simply as $\vsig\odot\vw\ge\0$ where the operator $\odot$ denotes the Hadamard product. 
The feasible region is expressed as
\begin{tsaligned}
  \cS :=
  \left\{
  \vw\in\bR^{d}\,\middle|\,
  \vsig\odot\vw\ge\0_{d}
  \right\}. 
\end{tsaligned}
The sign-constrained SVM learning problem is described as
\begin{tsaligned}\label{eq:scsvm-primal}
  \text{min}\quad
  &
  P(\vw)
  \quad
  \text{wrt}
  \quad
  \vw\in\cS,
\end{tsaligned}
where
\begin{tsaligned}\label{eq:scsvm-primalobj-def}
  &
  P(\vw)
  :=
  \frac{\lambda}{2}\lVert\vw\rVert^{2}
  +
  \frac{1}{n}
  \sum_{i=1}^{n}
  \max(0,1-y_{i}\left<\x_{i},\vw\right>). 
\end{tsaligned}
It can be seen that the optimization problem~\eqref{eq:scsvm-primal}
is the standard SVM learning problem if $\vsig=\0_{d}$.
In this study, two algorithms, a projected gradient algorithm and a Frank-Wolfe algorithm, were developed for solving the learning problem~\eqref{eq:scsvm-primal}. The two algorithms are presented in Section \ref{s:pega} and \ref{s:stdfw}, respectively.  

We shall use $\vw_{\star}$ to denote the solution optimal to our learning problem~\eqref{eq:scsvm-primal}.
A solution $\vw$ is said to be $\epsilon$-\emph{accurate} when $P(\vw)-P(\vw_{\star})\le\epsilon$.

\section{Projected Gradient Algorithm}
\label{s:pega}
Our projected gradient algorithm for the sign constrained SVM was developed by inserting the projection step into each step of Pegasos method~\cite{Shalev-Shwartz11a-pegasos}. 
The convergence theory of the original Pegasos method is based on the property that the optimal solution to the standard SVM learning problem is in the ball with radius $\sqrt{1/\lambda}$. Imposing the sign constraints breaks down this property, although we found that, if the sign constraints are imposed, the norm of the optimal solution is still bounded. The optimal solution lies in a slightly larger ball with radius $\sqrt{2/\lambda}$, denoted by $\cB:=\left\{ \vw\in\bR^{d}\,\middle|\,\lVert\vw\rVert\le\sqrt{2/\lambda}\right\}$. 
The projected gradient algorithm developed in this study starts with $\vw^{(1)}:=\0$. At the $t$-th iteration, the solution is updated as
\begin{tsaligned}\label{eq:scpega-update}
  \vw^{(t+1)}:=
  \Pi_{\cB}\left(\Pi_{\cS}\left(\vw^{(t)}-\frac{1}{\lambda t}\nabla P(\vw^{(t)})
  \right)\right). 
\end{tsaligned}
Therein, $\Pi_{\cB}(\x)$ and $\Pi_{\cS}(\x)$ denote the Euclidean projection from a point $\x\in\bR^{d}$ onto the ball $\cB$ and the feasible region $\cS\subseteq\bR^{d}$, respectively. 
The projection onto $\cS$ can be expressed as
\begin{tsaligned}
  \Pi_{\cS}(\vv) := \argmin_{\vw\in\cS}\lVert\vv-\vw\rVert
  = \vv + \max(\0,-\vsig\odot\vv).  
\end{tsaligned}
Assume that $\forall i\in[n]$, $\lVert\x_{i}\rVert\le R$. 
To analyze the convergence of the algorithm \eqref{eq:scpega-update},
we observe the following properties:
\begin{itemize}
\item $\cB\cap\cS$ is a closed convex set;
\item $\vw_{\star}\in\cB\cap\cS$;
\item
  $\forall\vw\in\bR^{d}$; 
  $\Pi_{\cB}\left(\Pi_{\cS}\left(\vw\right)\right)=\Pi_{\cB\cap\cS}\left(\vw\right)$;
\item  
  $\forall t\in\bN$,
  $\lVert\nabla P(\vw_{t})\rVert \le \sqrt{2\lambda} + R$.
\end{itemize}
Combining these properties with the proof techniques
used by Shalev-Shwartz et al~\cite{Shalev-Shwartz11a-pegasos}, 
we obtain the following theorem.
\begin{theorem-waku}
  It then holds that, $\forall T\in\bN$, $\exists T'\in[T]$,
  the primal objective error is bounded as 
  \begin{tsaligned}
    P(\vw^{(T')})-P(\vw_{\star}) \le
    \frac{(\sqrt{2\lambda}+R)^{2}\log(T)}{\lambda T}. 
  \end{tsaligned}
\end{theorem-waku}
A weak point of the projected gradient method is the lack of a termination criterion. There is no way to check the $\epsilon$-accuracy, because the minimal value $P(\vw_{\star})$ is usually unknown and thereby the objective error $P(\vw^{(t)})-P(\vw_{\star})$ cannot be assessed. 

\begin{figure}[t!]
  \centering
  \begin{tabular}{ll}
    &
    \includegraphics[width=0.45\linewidth]{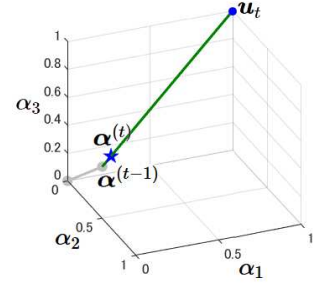}
  \end{tabular}
  \caption{%
    Iteration of Frank-Wolfe algorithm. In $t$-th iteration, the dual problem~\eqref{eq:prob-scsvm-dual} in which the dual objective $D$ is replaced to its linear approximation~\eqref{eq:lmo-scsvm} is solved. The solution is moved to the point at which the dual objective $D$ is maximized on the line segment between the current solution $\valph^{(t-1)}$ and the optimal solution to the sub-problem, say $\vu_{t}$.
    %
    \label{fig:demo894-06-fwiter}%
    }
\end{figure}
\begin{figure}[t!]
  \centering
  \begin{tabular}{lll}
    (a) &
    \\
    \multicolumn{2}{l}{%
      \includegraphics[width=0.73\linewidth]{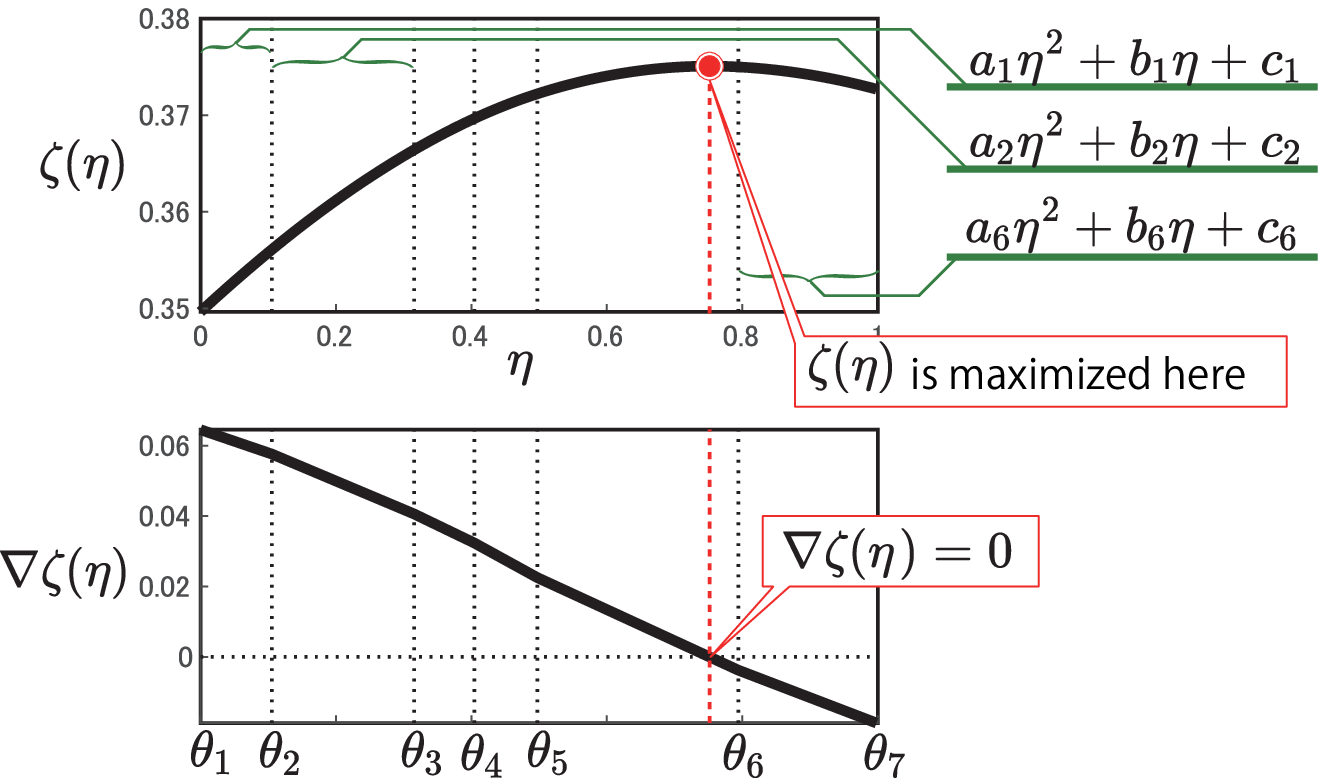}%
      }%
    \\
    (b) &
    (c) 
    \\
    \includegraphics[width=0.49\linewidth]{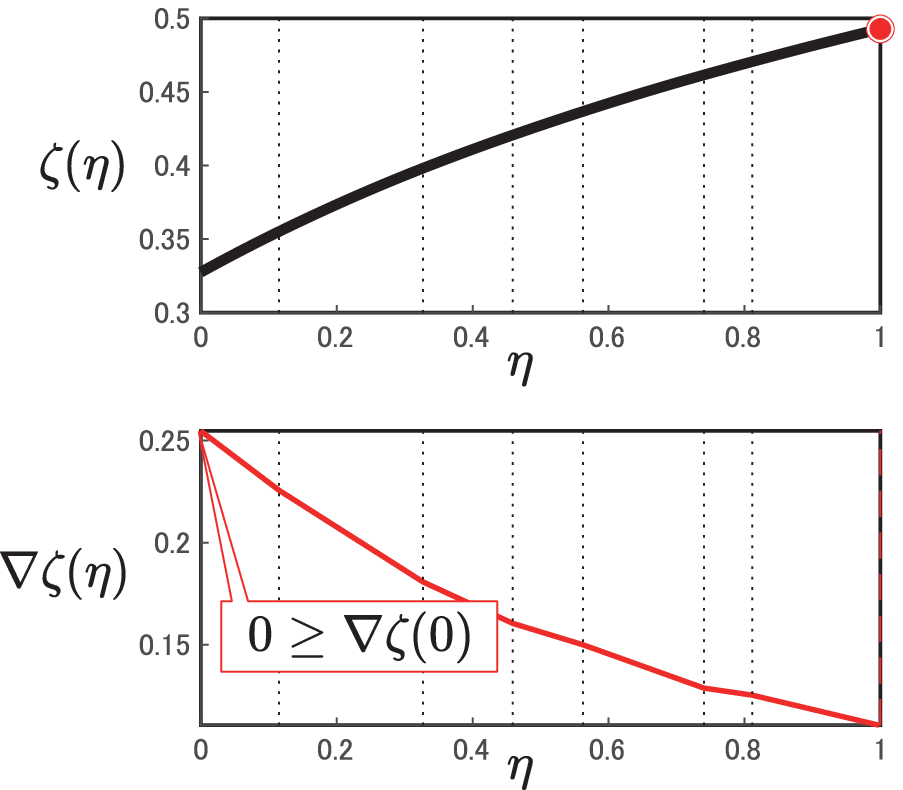}
    &
    \includegraphics[width=0.41\linewidth]{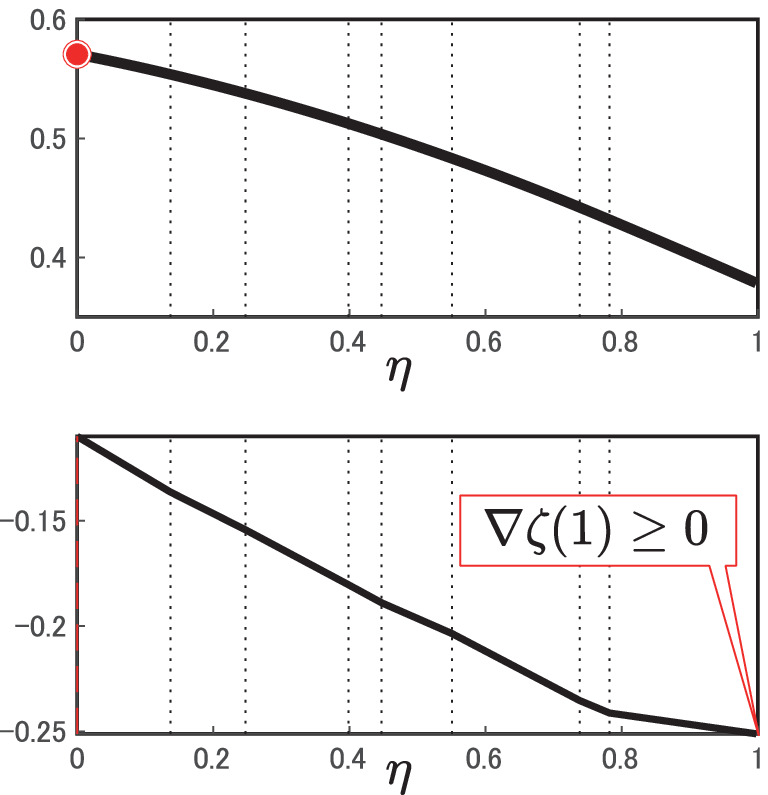}
  \end{tabular}
  \caption{
    Objective function for line search problem, say $\zeta(\eta)$, is a
    piecewise quadratic function.
    In the $h$-th interval $[\theta_{h},\theta_{h+1}]$,
    the function is expressed in the form of
    $\zeta(\eta)=a_{h}\eta^{2}+b_{h}\eta+c_{h}$. 
    There are three cases in the line search problem. 
    (a) One of three cases
    in which $\nabla\zeta(0)>0>\nabla\zeta(1)$.
    The point $\eta$ with the gradient vanishing
    is the maximizer of the function $\zeta(\eta)$
    in this case. 
    (b) In case of $0\ge\nabla\zeta(0)$,
    the function $\zeta(\eta)$ attains the maximum
    at $\eta=1$. 
    (c) In case of $\nabla\zeta(1)\ge 0$,
    $\zeta(\eta)$ is maximized at $\eta=0$. 
    \label{fig:demo890-01-zeta-3cases}
    }
\end{figure}
\section{Dual Problem}
\label{s:dual}
The projected gradient algorithm described in the previous section suffers from the lack of a clear criterion for when iterations should be terminated. 
To obtain a termination criterion, we consider solving a dual problem instead of the primal problem. The following problem is dual to the primal problem~\eqref{eq:scsvm-primal}: 
\begin{tsaligned}\label{eq:prob-scsvm-dual}
  \text{max}\quad&
  D(\valph)
  \quad
  \text{wrt}\quad
  \valph\in[0,1]^{n},
  \\
  \text{where}\quad&
  D(\valph):=
  -\frac{\lambda}{2}\lVert\vw(\valph)\rVert^{2} + \frac{1}{n}\left<\1,\valph\right>, 
  \\
  &
  \vw(\valph):=\Pi_{\cS}\left(\frac{1}{\lambda n}\X\valph\right). 
\end{tsaligned}
where a $d$-dimensional vector $y_{i}\x_{i}$ is stored in the $i$-th column of the matrix $\X\in\bR^{d\times n}$. 
The dual problem for the sign-constrained SVM is reduced to that of the standard SVM when no sign constraints are imposed, which can be seen as follows: in the case of no sign constraints, $\cS=\bR^{d}$, leading to $\Pi_{\cS}(\vv)=\vv$ and $\vw(\valph)=\X\valph/(\lambda n)$; substituting them into \eqref{eq:prob-scsvm-dual}, the dual problem for the standard SVM is derived.
This fact suggests that the difference in the dual objective functions between the sign-constrained and standard SVMs is the presence or absence of the projection onto the feasible region. The projection operation in the dual objective for the sign-constrained SVM precludes the direct use of the standard optimization approach to learning the standard SVM, and thereby makes the optimization problem challenging.

The primal variable optimal to the primal problem~\eqref{eq:scsvm-primal} can be recovered by $\vw_{\star}=\vw(\valph_{\star})$ where $\valph_{\star}$ is a solution optimal to the dual problem~\eqref{eq:prob-scsvm-dual}, which suggests that iterations in some dual optimization algorithm can be stopped when the duality gap is sufficiently small (i.e. $P(\vw(\valph))-D(\valph)\le\epsilon$). From the nature of the duality: 
\begin{tsaligned}
  \forall\valph\in[0,1]^{n},\qquad P(\vw(\valph))\ge P(\vw_{\star}) \ge D(\valph), 
\end{tsaligned}
the duality gap lower than a pre-defined threshold $\epsilon$ ensures the primal objective error below the threshold $\epsilon$. 

\section{Frank-Wolfe Algorithm}
\label{s:stdfw}
In this section, we present a dual optimization algorithm based on the Frank-Wolfe framework~\cite{Levitin66,jaggi13icml} for learning the sign constrained SVM. The Frank-Wolfe-based algorithm presented in this section is alternative to the projected gradient algorithm described in the previous section.

In general, the Frank-Wolfe framework performs optimization within a convex polyhedron. The feasible region of the dual problem for the sign-constrained SVM is a hyper-cube which is in a class of convex polyhedrons, suggesting that the dual problem can be a target of the Frank-Wolfe framework. 

\begin{algorithm}[t!]
  \caption{
    Frank-Wolfe algorithm for solving the dual problem \eqref{eq:prob-scsvm-dual}. 
    \label{algo:stdfw-svm}}
  \Begin{
    Let $\valph^{(0)}\in[0,1]^{n}$\;
    \For{$t:=1$ \KwTo $T$}{
      $\vu_{t}\in\argmax_{\vu\in[0,1]^{n}}\left<\nabla D(\valph^{(t-1)}),\vu\right>$\;    
    $\vq_{t}:=\vu_{t}-\valph^{(t-1)}$\;
    $\eta_{t}\in\arg\max_{\eta\in[0,1]}D(\valph^{(t-1)}+\eta\vq_{t})$\;
    $\valph^{(t)} := \valph^{(t-1)}+\eta_{t}\vq_{t}$\;
    }
  }
\end{algorithm}
As described in Algorithm~\ref{algo:stdfw-svm}, each iteration of the Frank-Wolfe framework consists of two steps, respectively, referred to as the \emph{direction finding step} and the \emph{line search step}.
In the direction finding step, a linear programming problem is solved.
In the sub-problem, the dual objective is replaced with its linear approximation
around a previous solution:
\begin{tsaligned}
  \vu\mapsto
  \left<\nabla D(\valph^{(t-1)}),\vu\right>
  +
  D(\valph^{(t-1)}). 
\end{tsaligned}
The linear approximation is maximized within the convex polyhedron, yielding
a linear programming problem. We denote by $\vu_{t}$
the optimal solution to the sub-problem at $t$-th iteration.

In the line search step, the dual objective is maximized on the line segment between the previous solution and the solution optimal to the linear programming problem, say $\valph^{(t-1)}$ and $\vu_{t}$, respectively.  The updated solution can be expressed as $\valph^{(t)}:=\valph^{(t-1)}+\eta_{t}\vq_{t}$ where
\begin{tsaligned}
  \eta_{t}\in
  \argmax_{\eta\in[0,1]}
  D(\valph^{(t-1)}+\eta\vq_{t})
  \text{ and }
  \vq_{t}:=\vu_{t}-\valph^{(t-1)}. 
\end{tsaligned}
As long as both the sub-problems involved in the two steps are exactly solved at each iteration,
the sublinear convergence is ensured. 
In addition, the accuracy of the solution is guaranteed by use of the duality gap $P(\vw(\valph))-D(\valph)$ for a stopping criterion. If the computational cost of each iteration in the Frank-Wolfe algorithm is within $O(nd)$, then the new approach share all the favorable properties of the projected gradient method and simultaneously overcomes the shortcoming.
These discussions suggest that 
both the direction finding step and the line search step need to be computed within $O(nd)$, in order to keep the computational cost of each iteration from exceeding $O(nd)$.

\subsection{Direction Finding Step}
\label{ss:lmo}
It is shown here that $O(nd)$ computation accomplishes the direction finding step in the Frank-Wolfe algorithm (Algorithm~\ref{algo:stdfw-svm}). The direction finding step requires to solve the following sub-problem
\begin{tsaligned}\label{eq:lmo-scsvm}
  \text{min}\quad&
  \left<\nabla D(\valph^{(t-1)}), \vu\right>
  \quad
  \text{wrt}\quad
  \vu\in[0,1]^{n}. 
\end{tsaligned}
The sub-problem~\eqref{eq:lmo-scsvm} is in a class of linear programming problems.
The direction finding step consumes $O(n^{3})$ computation if we resort to a general-purpose linear programming solver. The computational cost $O(n^{3})$ is computationally taxing when training with a large set of training examples. 

In this study, we have found a closed-form solution optimal to the linear programming problem~\eqref{eq:lmo-scsvm}. The $i$-th entry in the optimal solution $\vu_{t}\in[0,1]^{n}$ is expressed as
\begin{tsaligned}\label{eq:lmosol-scsvm}
  u_{i,t}
  =
  \begin{cases}
    1 \qquad \text{if }y_{i}\left<\vw(\valph^{(t-1)}),\x_{i}\right> < 1,
    \\
    0 \qquad \text{if }y_{i}\left<\vw(\valph^{(t-1)}),\x_{i}\right> \ge 1. 
  \end{cases}
\end{tsaligned}
\tslong{See Subsection~\ref{ss:deriv-eq:lmosol-scsvm}
for the derivation. }%
\tsshort{See the supplement
for the derivation. }%
The optimal solutions to the sub-problem~\eqref{eq:lmo-scsvm} is not unique.
The sub-linear convergence is ensured even if taking any point among the set of the optimal solutions. 

\textbf{Time Complexity of Direction Finding Step: }
The procedure and the time complexities for computation of the closed-form solution \eqref{eq:lmosol-scsvm}
are summarized as follows.
\begin{center}
  \begin{tabular}{ll}
    Compute $\vv^{(t-1)} := \X\valph^{(t-1)}/(\lambda n)$;
    & $O(nd)$.
    \\
    Compute $\vw^{(t-1)} := \Pi_{\cS}(\vv^{(t-1)})$; 
    & $O(d)$.
    \\
    Compute 
    $\z^{(t-1)} := \X^\top\vw^{(t-1)}$; 
    & $O(nd)$.
    \\
    $\forall i\in[n]$, 
    $ u_{i,t} := \mathds{1}(1\le z_{i}^{(t-1)})$; 
    & $O(n)$.   
  \end{tabular}
\end{center}
Therein, $z_{i}^{(t-1)}$ is the $i$-th entry in the vector $\z^{(t-1)}$, 
and the function $\mathds{1}(\cdot)$ takes the value of one if the logical argument is
true; otherwise zero. 
The line search problem can, thus, be solved in $O(nd)$
when $d$ is within $O(n)$. 

\subsection{Line Search Step}
\label{ss:lnsrch}
It has been seen that the direction finding step can be performed within $O(nd)$ computational cost. It shall be shown here that the line search step requires only $O(nd)$ cost, too.
Line search is an operation that finds an optimal solution, denoted by $\eta_{\star}\in[0,1]$,
to the sub-problem: 
\begin{tsaligned}\label{eq:prob-lnsrch}
  \text{max}\quad
  &
  \zeta(\eta)
  \quad
  \text{wrt}\quad
  \eta\in[0,1],
  \\
  \text{where}\quad
  &
  \zeta(\eta):=D(\valph+\eta\vq),
  \quad
  \valph \in [0,1]^{n},
  \\
  & \vq\in [0,1]^{n}-\valph,\,\, \vq\ne\0_{n}. 
\end{tsaligned}
In case of $\vq=\0_{n}$, it is obvious that any $\eta\in[0,1]$ is optimal. Here, our discussion is limited to the case $\vq\ne \0_{n}$ as given in \eqref{eq:prob-lnsrch}. 
The following is the key lemma for solving the line search problem. 
\begin{lemma-waku}\label{lem:lnsrch-is-piesequadfun}
  The objective function of the sub-problem given in \eqref{eq:prob-lnsrch},
  say $\zeta:[0,1]\to\bR$, is a differentiable and concave piecewise quadratic function
  and its derivative is continuous and monotonically decreasing. 
  Namely, there exist an integer $d_{t}\in[d+1]$, 
  coefficients $(a_{k},b_{k},c_{k})$ for $k\in[d_{t}]$
  with $a_{k}\le 0$ and
  $\theta_{0},\dots,\theta_{d_{t}+1}$ such that 
  $0 = \theta_{1} < \theta_{2} < \dots < \theta_{d_{t}+1} = 1$ 
  and the function can be expressed as 
  $\forall k\in[d_{t}]$,
  $\forall \eta\in[\theta_{k},\theta_{k+1}]$, 
  \begin{tsaligned}\label{eq:01-lnsrch-is-piesequadfun}
    \zeta(\eta) = a_{k}\eta^{2} + b_{k}\eta + c_{k},  
  \end{tsaligned}
  and $\forall k\in[d_{t}-1]$,
  \begin{tsaligned}\label{eq:02-lnsrch-is-piesequadfun}
    2 a_{k}\theta_{k+1} + b_{k} = 2 a_{k+1}\theta_{k+1} + b_{k+1}. 
  \end{tsaligned}
\end{lemma-waku}
\tslong{See Subsection~\ref{ss:proof-lem:lnsrch-is-piesequadfun}
for the proof of Lemma~\ref{lem:lnsrch-is-piesequadfun}. }%
\tsshort{See the supplement
for the proof of Lemma~\ref{lem:lnsrch-is-piesequadfun}. }%

\textbf{Solution to Line Search: }
Lemma~\ref{lem:lnsrch-is-piesequadfun} suggests that 
three cases described in Figure~\ref{fig:demo890-01-zeta-3cases}. 
From this observation, it turns out that 
an optimal solution to the line search problem \eqref{eq:prob-lnsrch} can be
given in a closed-form as
\begin{tsaligned}\label{eq:sol-to-lnsrch}
  \eta_{\star}
  =
  \begin{cases}
    0 \qquad& \text{if }b_{1} \le 0,
    \\
    1 \qquad& \text{if }2 a_{d_{t}+1} + b_{d_{t}+1} \ge 0,
    \\
    -
    \frac{b_{k_{\star}}}{2a_{k_{\star}}}
    \qquad&
    \text{if $b_{1} > 0$,  $2 a_{d_{t}+1} + b_{d_{t}+1} < 0$, $a_{k_{\star}}< 0$,}
    \\
    \theta_{k_{\star}}
    \qquad&
    \text{if $b_{1} > 0$,  $2 a_{d_{t}+1} + b_{d_{t}+1} < 0$, $a_{k_{\star}}= 0$}
  \end{cases}
\end{tsaligned}
where $k_{\star}\in[d_{t}]$ here is the index of one of
$d_{t}$ intervals, 
in which the derivative vanishes at some point. Namely, 
\begin{tsaligned}
    2 a_{k_{\star}}\theta_{k_{\star}} + b_{k_{\star}} \ge 0 \quad\text{ and }\quad
    2 a_{k_{\star}}\theta_{k_{\star}+1} + b_{k_{\star}} \le 0. 
\end{tsaligned}
Note that, from the definition of $k_{\star}$, it must hold that $b_{k_{\star}}=0$
in the case
that $b_{1} > 0$,  $2 a_{d_{t}+1} + b_{d_{t}+1} < 0$ and $a_{k_{\star}}= 0$. 

\textbf{Endpoints and Coefficients of Piecewise Quadratic Function $\zeta$: }
Here we describe
how to determine the endpoints
$\theta_{1},\theta_{2},\dots,\theta_{d_{t}+1}\in[0,1]$
and the coefficients of the piecewise quadratic function $\zeta$,
say $(a_{k},b_{k},c_{k})$ for $k=1,\dots,d_{t}$. 

Let $\vv_{0}:=\left[v_{1,0},\dots,v_{d,0}\right]^{\top}$
and $\vv_{q}:=\left[v_{1,q},\dots,v_{d,q}\right]^{\top}$
have the following entries, $\forall h\in[d]$, 
\begin{tsaligned}
  v_{h,0} := \frac{1}{\lambda n}\left<\vf_{h},\valph\right>, \quad
  v_{h,q} := \frac{1}{\lambda n}\left<\vf_{h},\vu-\valph\right>. 
\end{tsaligned}
where $\vf_{h}\in\bR^{n}$ is the $h$-th column vector in $\X^{\top}$.
Let
\begin{multline}
  \Theta
  :=  
  \Bigg\{
  \theta\in[0,1]
  \,\Bigg|\,
  \exists h\in\cI_{+}
  \\
  \text{ s.t. }
  v_{h,q} \ne 0, \,
  \theta = - \frac{v_{h,0}}{v_{h,q}}
  \Bigg\}
  \cup \{ 0, 1\}. 
\end{multline}
The number of intervals $d_{t}$ can be set to
the cardinality of the set $\Theta$ minus one
(i.e. $d_{t} = \text{card}(\Theta)-1$)
and the endpoints $\theta_{1},\dots,\theta_{d_{t}+1}$ of $d_{t}$
intervals are determined by sorting the elements in $\Theta$
so that $\theta_{1} < \theta_{2} < \dots < \theta_{d_{t}+1}$. 
In this setting, it can be observed that 
equation~\eqref{eq:01-lnsrch-is-piesequadfun} is satisifed
if coefficients are given as
$\forall k\in[d_{t}]$,
\begin{tsaligned}\label{eq:coef-piesequadfun}
  & a_{k}:=- \frac{\lambda}{2}
  \sum_{h\in\cH_{k}} v_{h,q}^{2},
  \\
  &
  b_{k}:=
  \frac{1}{n}\left<\1,\vu-\valph\right>
  -
  \lambda \sum_{h\in\cH_{k}} v_{h,q}v_{h,0},
  \\
  &
  c_{k}
  :=
  \frac{1}{n}\left<\1,\valph\right> 
  -
  \frac{\lambda}{2}
  \sum_{h\in\cH_{k}} v_{h,0}^{2}
\end{tsaligned}
where
\begin{tsaligned}
  \cH_{k}
  :=
  \cI_{0}
  \cup
  \left\{
  h\in\cI_{+}
  \,\middle|\,
  v_{h,0} + \frac{\theta_{k}+\theta_{k+1}}{2} v_{h,q} > 0
  \right\}. 
\end{tsaligned}

\textbf{Time Complexity of Line Search Step: }
The procedure required to compute the solution \eqref{eq:sol-to-lnsrch}
and the time complexity of each step is described as follows. 
\begin{center}
  \begin{tabular}{ll}
    Compute $\vv_{0}$ and $\vv_{q}$; & $O(nd)$. 
    \\
    Determine $\Theta$; & $O(d)$. 
    \\
    Sort the elements in $\Theta$; & $O(d\log d)$. 
    \\
    Compute $\cH_{k}$ for $k\in[d_{t}]$; & $O(d^{2})$. 
    \\
    Compute coefficients $(a_{k},b_{k},c_{k})$ 
    for $k\in[d_{t}]$; & $O(d^{2})$. 
    \\
    Find $k_{\star}$; & $O(d)$.  
    \\
    Compute a solution \eqref{eq:sol-to-lnsrch};
    & $O(1)$.  
  \end{tabular}
\end{center}
Hence the line search problem can be solved in $O(nd)$
when $d$ is within $O(n)$. 
\subsection{Convergence Analysis}
\label{ss:fwconv}
We have seen that the direction finding step and
the line search step can be performed exactly, 
meaning that our algorithm inherits the theoretical guarantee
for the convergence property of the Frank-Wolfe algorithm. 
In general, the objective error of the Frank-Wolfe algorithm
for a minimization problem is bounded as
\begin{tsaligned}
  F(\x^{(T)}) - F(\x_{\star})
  \le
  \frac{2C_{F}}{T+2}
\end{tsaligned}
where $C_{F}$ is the curvature of the
continuously differentiable and convex objective function $F$
to be minimized~\cite{jaggi13icml}.
The curvature~\cite{jaggi13icml} is defined as
\begin{tsaligned}\label{eq:curv-def}
  C_{F} :=
  \sup
  \Big\{
  \frac{2}{\gamma^{2}}\cD_{F}(\y_{\gamma}\,;\,\x)
  \,\Big| & \,
  \vu, \x\in\cM,\, \gamma\in(0,1],\,
    \\
    &
  \y_{\gamma}:=(1-\gamma)\x + \gamma\vu 
  \Big\},  
\end{tsaligned}
where $\cM$ is the domain of the minimization problem
and $\cD_{F}(\y\,;\,\x)$ is the Bregman divergence.
We obtained the following bound for the curvature of
the objective in the dual problem~\eqref{eq:prob-scsvm-dual}.
\begin{lemma-waku}\label{lem:curv-scsvm}
  Assume that the norm of every feature vector is
  $R$, at most (i.e. $\forall i\in[n]$, $\lVert\x_{i}\rVert\le R$). 
  The curvature for $F:=-D$ with $\cM:=[0,1]^{n}$
  is then bounded as $C_{F}\le R^{2}/\lambda$. 
\end{lemma-waku}
\tslong{See Subsection~\ref{ss:proof-lem:curv-scsvm}
for the proof of Lemma~\ref{lem:curv-scsvm}. }%
\tsshort{See the supplement
for the proof of Lemma~\ref{lem:curv-scsvm}. }%

This lemma leads to our main result: 
\begin{theorem-waku}
  Suppose that $\forall i\in[n]$, $\lVert\x_{i}\rVert\le R$. 
  Each iteration of Algorithm~\ref{algo:stdfw-svm} can be
  performed within $O(nd)$ computational cost.
  The objective error is less than or equal to a positive scalar $\epsilon>0$
  (i.e. $D(\valph_{\star}) - D(\valph^{(T)}) \le \epsilon$) for any $T\in\bN$
  such that 
  \begin{tsaligned}
    T \ge \frac{2R^{2}}{\lambda \epsilon} - 2.     
  \end{tsaligned}
\end{theorem-waku}
\begin{table}[t!]
  \centering
  \caption{ROC scores for amino acid sequence classification.
    \label{tab:demo387-06}}
  {
  \begin{tabular}{|c|cc|}
    \hline
    Class
    & \shortstack{Conventional \\ SVM-pairwise}
    & \shortstack{Sign Constrained \\ SVM-pairwise}
    \\
    \hline    
1 & 0.731 (0.009)  & \textbf{\underbar{0.749}} (0.010) \\
2 & 0.681 (0.013)  & \textbf{\underbar{0.756}} (0.012) \\
3 & 0.735 (0.013)  & \textbf{\underbar{0.758}} (0.012) \\
4 & 0.745 (0.010)  & \textbf{\underbar{0.768}} (0.009) \\
5 & 0.709 (0.017)  & \textbf{\underbar{0.784}} (0.008) \\
6 & 0.625 (0.007)  & \textbf{\underbar{0.692}} (0.012) \\
7 & 0.628 (0.023)  & \textbf{\underbar{0.702}} (0.030) \\
8 & 0.664 (0.018)  & \textbf{\underbar{0.733}} (0.018) \\
9 & 0.603 (0.019)  & \textbf{\underbar{0.681}} (0.022) \\
10 & 0.712 (0.008)  & \textbf{\underbar{0.739}} (0.010) \\
11 & 0.523 (0.026)  & \textbf{\underbar{0.561}} (0.029) \\
12 & 0.894 (0.013)  & \textbf{\underbar{0.905}} (0.012) \\
    \hline
  \end{tabular}
  }
\end{table}
\section{Experiments}
\label{s:exp}
In this section, we illustrate how fast the optimization algorithm converge in real-world data, and empirically show how well the sign constraints work for the task of biological sequence classification. 
\subsection{Convergence}
We examine the convergence of the projected gradient and the Frank-Wolfe algorithms on the Cora and MNIST datasets.
Cora dataset is a collection of 15,396 academic papers related to computer science~\cite{SenNamBil08}. Each paper is annotated manually to provide a hierarchical classification. The dataset has ten categories at the top hierarchy. We divided the top categories into two groups to pose a binary classification problem. The positive class consists of IR, AI, OS, HA, and Pr, and the negative class is from Da, EC, Ne, DA, and HCI, resulting 10,778 and 4,618 papers belong to the positive and negative class, respectively. The task here is to classify each paper from the normalized word frequencies in its title. After removing stop words, 12,644 words used in the title of 15,396 papers were found. Each 12,644-dimensional feature vector is normalized with Euclidean norm. 
The MNIST dataset is the most popular dataset for benchmarking machine learning methodologies. The dataset has 60,000 handwritten digit images. Each image has 784 pixels. For this simulation, images of odd digits (i.e. `1', `3', `5', `7', `9') and even digits (i.e. `0', `2', `4', `6', `8') were labeled as positive and negative, respectively. Each feature vector is normalized to transform each of examples to a unit vector.

A hundred iterations of the projected gradient and the Frank-Wolfe algorithms were run and the primal objective values $P(\vw)$ were monitored 55 times at different iterations. The iteration numbers at which the objective function was evaluated were spaced evenly on a logarithmic scale (i.e. $t=1,2,3,\dots,50,53,55,\dots,87,92,96,100$). The minima among these objective errors monitored so far were recorded. For the Frank-Wolfe Algorithm, the duality gaps $P(\vw)-D(\valph)$ were also evaluated.  The regularization parameter $\lambda$ was varied with $\lambda=10^{-6}/n, 10^{-4}/n, 10^{-2}/n$ where $n=15,396$ for Cora and $n=60,000$ for MNIST.
In order to examine how fast each algorithm converges, the true objective error should be assesed, although it is impossible because $P(\vw_{\star})$ is usually unknown. To approximate the objective error, we ran the Frank-Wolfe algorithm to compute the dual objective value at $T':=1,000$th iteration. Although $D(\valph^{(T')})$ might be slightly smaller than the true minimal primal objective value $\min_{\vw\in\cS}P(\vw)$, the value $P(\vw)-D(\valph^{(T')})$ was regarded as the objective error.

\tsshort{%
The upper row in Figure~\REFFIGDEMOCORA~of the supplement plots the objective errors against iteration numbers on Cora, respectively. 
}%
\tslong{%
The upper row in Figure~\REFFIGDEMOCORA~plots the objective errors against iteration numbers on Cora, respectively. 
}
For all $\lambda$ on Cora, the projected gradient algorithm achieved smaller objective errors at around ten iterations, whereas the objective errors of the Frank-Wolfe were much smaller than those of the projected gradient after 15 iterations.
\tsshort{%
For MNIST (See Figure~\REFFIGDEMOMNIST~in the supplement), 
}%
\tslong{%
For MNIST (See Figure~\REFFIGDEMOMNIST), 
}%
the Frank-Wolfe converged to the minimum much faster than the projected gradient. The lower row depicts the duality gaps produced from the Frank-Wolfe.  Recall that the duality gap can be a clear stopping criterion ensuring the quality of the solution. 
The duality gaps were indeed converged to zero, implying that the duality gap can be utilized as a stopping criterion.  

\subsection{Biological Sequence Classification}
\label{ss:svmpairwise}
Here we demonstrate that sign constraints are a powerful methodology for boosting the prediction performance for SVM-pairwise~\cite{LiaNob03-jcb} which is a binary classifier of biological sequences. SVM-pairwise predicts the existence/absence of a property of a query protein from its amino-acid sequence.
Conventional analysis of biological sequences is based on pairwise comparison between two sequences using a sequence similarity measure such as the Smith Waterman score~\cite{SmiWat81}.
SVM-pairwise was developed to predict the remote homology by combining the sequence comparison with supervised machine learning. The feature vector taken by SVM-pairwise is the sequence similarities to each of sequences for training. If $n$ proteins are in a training dataset, the feature vector has $n$ entries, $x_{1},\dots,x_{n}$. If we assume that the first $n_{+}$ proteins in the training set are annotated as positive and the rest of $n_{-}:=n-n_{+}$ proteins negative, then the first $n_{+}$ entries $x_{1},\dots,x_{n_+}$ are the sequence similarities to positive proteins and $x_{n_{+}+1},\dots,x_{n}$ are the similarities to negative proteins. Since each feature vector for training is $n$-dimensional, the data matrix $\X$ is square 
\tsshort{%
(See Figure~\REFFIGSVMPAIRWISE~in the supplement). 
}%
\tslong{%
(See Figure~\REFFIGSVMPAIRWISE). 
}%
The prediction score for a query protein $\x$ is the difference from
\begin{tsaligned}
  w_{1}x_{1} + \dots + w_{n_{+}}x_{n_{+}}
\end{tsaligned}
to
\begin{tsaligned}
  - w_{n_{+}+1}x_{n_{+}+1} - \dots - w_{n}x_{n}. 
\end{tsaligned}
%
From this view, it is preferable that the first $n_{+}$ weight coefficients $w_{1},\dots,w_{n_{+}}$ are non-negative and that the remaining $n_{-}$ coefficients $w_{ n_{+}+1},\dots,w_{n}$ are non-positive. Nevertheless, the pairwise SVM does not ensure these requirements.
These observations motivated us to impose these constraints explicitly as
$w_{1} \ge 0, \dots, w_{n_{+}}\ge 0$
and
$w_{n_{+}+1}\le 0, \dots, w_{n}\le 0$.

We examined the effectiveness of sign constraints on 3,583 yeast proteins with the annotations for these cellular functions and the Smith-Waterman similarities available from https://noble.gs.washington.edu/proj/sdp-svm/. The annotation of cellular function of a protein was existence or absence for each of 12 functional categories. We posed 12 independent binary classification task. A half of 3,583 proteins were selected randomly and used for training and the rest were used for testing.
We performed five-fold cross-validation within a training subset to determine the value of the regularization constant.
Weight coefficients $\vw\in\bR^{n}$ were determined by conventional SVM-pairwise and sign-constrained SVM-pairwise, respectively. For each of 12 binary classification tasks, ten ROC scores for conventional SVM-pairwise and ten ROC scores for sign-constrained SVM-pairwise were obtained by repeating this procedure ten times.
The one-sample t-test was performed to detect the significant difference between two ROC scores. 
The average of ROC scores and the standard deviations over ten repetitions were shown in Table~\ref{tab:demo387-06}.
The best ROC scores are bold-faced and the underlined scores do not have a statistical significant difference from the best performance. 
\tsshort{%
  Additional results are given in the supplement.
}%
\tslong{%
  Additional results are given in Subsection~\ref{ss:protein}. 
}%
For all of 12 tasks, sign constrained modification consistently improved the ROC scores, suggesting that the sign constraints are a promising technique for SVM-pairwise framework.

\section{Conclusion}
\label{s:concl}

In this paper, we discussed the projected gradient algorithm and the Frank-Wolfe algorithm for learning the linear SVM under sign constraints. We provided solid analyses for the computational cost of each iteration and the convergence rate both for the two optimization algorithms. We found that both algorithms were ensured to have the sublinear convergence rate and the $O(nd)$ computational cost for each iteration. Finally, we proposed to introduce sign constraints for the SVM-pairwise approach and empirically demonstrated the promising prediction performances on 12 binary classification tasks for cellular functional prediction. SVM-pairwise is an approach specialized for computational biology, although similar approaches have also been discussed in more general settings~\cite{Tsuda02esann,Ng01nips}. Benchmarking the effects of sign constraints on SVM-pairwise in other applications will be the subject of future work.

\section*{Acknowledgment}
This research was performed by the Environment Research and Technology Development Fund JPMEERF20205006 of the Environmental Restoration and Conservation Agency of Japan and supported partially by JSPS KAKENHI Grant Number 19K04661. 

\bibliographystyle{plain}

\newpage
\section{Derivations and Proofs}
\subsection{Derivation for \eqref{eq:lmosol-scsvm}}
\label{ss:deriv-eq:lmosol-scsvm}
Observe that the linear programming problem of
an $n$-dimensional variable $\vu$ given in \eqref{eq:lmo-scsvm}
can be decomposed into $n$ independent problems: $\forall i\in[n]$, 
\begin{tsaligned}
  \max\quad & g_{i}^{(t-1)}u_{i}\quad \text{wrt}\quad u_{i}\in[0,1],
  \\
  \text{where}\quad & g_{i}^{(t-1)}
  := \left.\frac{\partial D(\valph)}{\partial \alpha_{i}}\right|_{\valph=\valph^{(t-1)}}. 
\end{tsaligned}
It can be seen that the gradient $g_{i}^{(t-1)}$ is expressed as
\begin{tsaligned}
  g_{i}^{(t-1)} = \frac{1}{n}
  \left(1-\left<\vw(\valph^{(t-1)}),\x_{i}\right>\right)
\end{tsaligned}
From this, we get
\begin{multline}
  \mathds{1}\left[ \left<\vw(\valph^{(t-1)}),\x_{i}\right> < 1 \right]
  =
  \\
  \mathds{1}\left[ g_{i}^{(t-1)} > 0 \right]  
  \in \text{argmax}_{u_{i}\in[0,1]}g_{i}^{(t-1)}u_{i}
\end{multline}
which implies that the vector with $i$-th entry
$\mathds{1}\left[ \left<\vw(\valph^{(t-1)}),\x_{i}\right> < 1 \right]$, 
as defined in \eqref{eq:lmosol-scsvm},
is a solution optimal to the linear programming problem~\eqref{eq:lmo-scsvm}. 
\qed

\subsection{Proof for Lemma~\ref{lem:lnsrch-is-piesequadfun}}
\label{ss:proof-lem:lnsrch-is-piesequadfun}
The fact that $\forall k\in[d_{t}]$, $a_{k}\le 0$ is apparent when
coefficients and endpoints are determined as 
in \eqref{eq:coef-piesequadfun} and the descriptions around the equation.
Let $w_{h}(\valph)$ and $v_{h}(\valph)$ be the $h$-th entries
in two vectors $\vw(\valph)$ and $\vv(\valph)$, respectively. 
Let $\vq:=\vu-\valph$. 
We first use the following lemma to show that
the setting of coefficients and endpoints defined in
Subsection~\ref{ss:lnsrch}
actually establishes \eqref{eq:01-lnsrch-is-piesequadfun}.
\begin{lemma-waku}\label{lem:wh-is-v-if-k-in-Hk}
  Let an arbitrary $\eta\in[0,1]$ and
  let $k\in[d_{t}]$ such that $\eta\in[\theta_{k},\theta_{k+1}]$.
  It then holds that 
\begin{tsaligned}\label{eq:wh-is-v-if-k-in-Hk}
  w_{h}(\valph+\eta\vq)
  =
  \begin{cases}
    v_{h,0} + \eta v_{h,q}
    &\qquad\text{if }h\in\cH_{k},
    \\
    0
    &\qquad\text{if }h\in[d]\setminus\cH_{k},    
  \end{cases}
\end{tsaligned}
\end{lemma-waku}

\textbf{\footnotesize Proof for Lemma~\ref{lem:wh-is-v-if-k-in-Hk}}
The equality \eqref{eq:wh-is-v-if-k-in-Hk} holds apparently
for $h\in\cH_{k}\setminus \cI_{+}$.
We observe that
\begin{tsaligned}
  \forall \eta\in (\theta_{k},\theta_{k+1}), \quad
  v_{h,0} + \eta v_{h,q} \ne 0. 
\end{tsaligned}
For $h\in[d]\setminus\cH_{k} = \cI_{+}\setminus\cH_{k}$, the continuity
of the linear function
$\eta\mapsto v_{h,q} + \eta v_{h,q}$ implies that
\begin{tsaligned}
  \forall \eta\in(\theta_{k},\theta_{k+1}), \quad
  0 > v_{h,0} + \eta v_{h,q}
\end{tsaligned}
implying $w_{h}(\valph+\eta\vq) = 0$. 
For $h\in\cI_{+}\cap\cH_{k}$, the continuity
of the linear function
$\eta\mapsto v_{h,q} + \eta v_{h,q}$ implies
\begin{tsaligned}
  \forall \eta\in(\theta_{k},\theta_{k+1}), \quad
  0 < v_{h,0} + \eta v_{h,q}
  = w_{h}(\valph+\eta\vq). 
\end{tsaligned}
\begin{flushright}
\textbf{\footnotesize End of proof for Lemma~\ref{lem:wh-is-v-if-k-in-Hk}}
\end{flushright}

With help of Lemma~\ref{lem:wh-is-v-if-k-in-Hk},
the square of the norm of
the vector $\vw(\valph + \eta\vq)$
is expressed as
\begin{multline}
  \frac{1}{2}\lVert\vw(\valph + \eta\vq)\rVert^{2}
  =
  \frac{1}{2}\sum_{h\in[d]}(w_{h}(\valph + \eta\vq))^{2}
  \\
  =
  \frac{1}{2}\sum_{h\in\cH_{k}}
  v_{h,q}^{2}\eta^{2}
  +
  2 \eta v_{h,0}v_{h,q}
  +
  v_{h,0}^{2} 
\end{multline}
which derives the equality of
\eqref{eq:01-lnsrch-is-piesequadfun}
as
\begin{tsaligned}
  \zeta(\eta) &= D(\valph + \eta\vq)
  \\
  &=
  -\frac{\lambda}{2}\lVert\vw(\valph + \eta\vq)\rVert^{2}
  + \frac{1}{n}\left<\1,\valph + \eta\vq\right>
  \\
  &=
  -
  \frac{\lambda}{2}
  \sum_{h\in\cH_{k}}
  v_{h,q}^{2}\eta^{2}
  +
  \left(
  \frac{1}{n}\left<\1,\vq\right>
  -
  \lambda
  \sum_{h\in\cH_{k}}
  v_{h,0}v_{h,q}
  \right)
  \eta
  \\
  &\qquad +
  \frac{1}{n}\left<\1,\valph\right>
  -
  \frac{\lambda}{2}
  \sum_{h\in\cH_{k}}
  v_{h,0}^{2}
  \\
  &=
  a_{k}\eta^{2} + b_{k}\eta + c_{k}. 
\end{tsaligned}

We next show the equality of \eqref{eq:02-lnsrch-is-piesequadfun}.
We partition the index set $\cH_{k}$ into three exclusive subsets
$\cH_{1,k}$, $\cH_{2,k}$, $\cH_{3,k}$, each of which is defined as
\begin{tsaligned}
  & \cH_{1,k} := \cH_{k}\cap\cH_{k+1},
  &
  & \cH_{2,k} := \cH_{k}\setminus\cH_{k+1},
  \\
  & \cH_{3,k} := \cH_{k+1}\setminus\cH_{k},  
\end{tsaligned}
respectively.
Observe that 
\begin{tsaligned}\label{eq:28-lem-lnsrch-is-piesequadfun}
  \forall h\in\cH_{2,k}\cup\cH_{3,k},
  \qquad
  v_{h,0} + \theta_{k+1}v_{h,q} = 0, 
\end{tsaligned}
which follows from the combination of two
properties: the continuity of the linear function
$\eta\mapsto v_{h,q} + \eta v_{h,q}$ and
the change of the sign at $\eta=\theta_{k+1}$, i.e., 
\begin{tsaligned}
  \forall h\in\cH_{2,k}, 
  \quad
  & \forall \eta_{1}\in(\theta_{k},\theta_{k+1}), \quad
  v_{h,0} + \eta_{1}v_{h,q} > 0,
  \text{ and }
  \\
  & \forall \eta_{2}\in(\theta_{k+1},\theta_{k+2}), \quad
  v_{h,0} + \eta_{2}v_{h,q} \le 0, 
\end{tsaligned}
and
\begin{tsaligned}
  \forall h\in\cH_{3,k}, 
  \quad
  & \forall \eta_{1}\in(\theta_{k},\theta_{k+1}), \quad
  v_{h,0} + \eta_{1}v_{h,q} \le 0,
  \text{ and }
  \\
  & \forall \eta_{2}\in(\theta_{k+1},\theta_{k+2}), \quad
  v_{h,0} + \eta_{2}v_{h,q} > 0. 
\end{tsaligned}
From \eqref{eq:28-lem-lnsrch-is-piesequadfun}, 
the difference of RHS from LHS in \eqref{eq:02-lnsrch-is-piesequadfun}
can be written as
\begin{tsaligned}
  & 2 a_{k+1}\theta_{k+1}+b_{k+1} - (2a_{k}\theta_{k+1}+b_{k})
  \\
  & =
  \sum_{h\in\cH_{2,k}}
  \left( v_{h,q}^{2}\theta_{k+1} + v_{h,q}v_{h,0} \right)
  \\
  &\qquad -
  \sum_{h\in\cH_{3,k}}
  \left( v_{h,q}^{2}\theta_{k+1} + v_{h,q}v_{h,0} \right)
  \\
  &=
  \sum_{h\in\cH_{2,k}}
  \left( v_{h,q}\theta_{k+1} + v_{h,0} \right)v_{h,q}
  \\
  &\qquad -
  \sum_{h\in\cH_{3,k}}
  \left( v_{h,q}\theta_{k+1} + v_{h,0} \right)v_{h,q}
  \\
  &=
  \sum_{h\in\cH_{2,k}}
  0
  -
  \sum_{h\in\cH_{3,k}}
  0
  = 0. 
\end{tsaligned}
\qed

\subsection{Proof for Lemma~\ref{lem:curv-scsvm}}
\label{ss:proof-lem:curv-scsvm}
Let $\vf_{h}$ be the $h$th column of $\X^\top$.
We then have
\begin{tsaligned}
  v_{h}(\valph) &= \frac{1}{\lambda n}\left<\vf_{h},\valph\right>
  \qquad \text{and}
  \\
  w_{h}(\valph) &=
  \begin{cases}
    v_{h}(\valph) &\quad\text{ if }\sigma_{h}=0,
    \\
    \max(0,v_{h}(\valph)) &\quad\text{ if }\sigma_{h}=1. 
  \end{cases}
\end{tsaligned}
Let
\begin{tsaligned}
  \cH_{\valph}
  :=
  \left\{ h\in[d]\,\middle|\,
  \sigma_{h}=0 \text{ or }
  v_{h}(\valph) > 0
  \right\}
\end{tsaligned}
and
\begin{tsaligned}
  \vK_{\valph}
  :=
  \sum_{h\in\cH_{\valph}}
  \vf_{h}\vf_{h}^\top. 
\end{tsaligned}
For the negative second-order derivative of the objective,
we have 
\begin{multline}\label{eq:neg-hess-le-kern}
  -\nabla^{2}D(\valph)
  =
  \frac{1}{\lambda n^{2}}\vK_{\valph}
  =
  \frac{1}{\lambda n^{2}}
  \sum_{h\in\cH_{\valph}}
  \vf_{h}\vf_{h}^\top
  \\
  \preceq
  \frac{1}{\lambda n^{2}}
  \sum_{h=1}^{d}
  \vf_{h}\vf_{h}^\top
  =
  \frac{1}{\lambda n^{2}}\X^\top\X. 
\end{multline}
Therein, we have used the notation $\preceq$ to denote
the generalized inequality associated with the positive
semidefinite cone (i.e. for two symmetric matrices, $\vA$ and $\vB$,
we can write $\vA\preceq\vB$ if $\vB-\vA$ is positive semidefinite. ). 

The term of Bregman divergence in the definition of
the curvature \eqref{eq:curv-def} is bounded as
$\forall\vu,\forall\valph\in\cM$,
$\forall\gamma\in[0,1]$,
$\exists\eta\in[0,\gamma]$,
\begin{tsaligned}
  \cD_{-D}(\y_{\gamma}\,;\,\valph)
  &=
  D(\valph)-D(\y_{\gamma})+\left<\nabla D(\valph),\y_{\gamma}-\valph)\right>
  \\
  &=
  -\frac{1}{2}
  \left<\y_{\gamma}-\valph,
  \nabla^{2}D(\y_{\eta})(\y_{\gamma}-\valph)\right>
  \\
  &=
  -\frac{\gamma^{2}}{2}
  \left<\vu-\valph,
  \nabla^{2}D(\y_{\eta})(\vu-\valph)\right>
  \\
  &\le
  \frac{\gamma^{2}}{2\lambda n^{2}}
  \left<\vu-\valph,
  \X^\top\X(\vu-\valph)\right>
  \\
  &=
  \frac{\gamma^{2}}{2\lambda n^{2}}
  \lVert \X (\vu-\valph)\rVert^{2}
\end{tsaligned}
where we have used Taylor's theorem to obtain
the equality in the second line, and
the inequality in the fourth line follows
from \eqref{eq:neg-hess-le-kern}. 
Therefore, we have
\begin{tsaligned}
  C_{F}
  &=
  \sup\Bigg\{
  \frac{2}{\gamma^{2}}\cD_{-D}(\y_{\gamma}\,;\,\valph)
  \,\Bigg|\,
  \\
  &\qquad\qquad
  \vu,\valph\in[0,1]^{n},\, \gamma\in(0,1], \eta\in[0,\gamma]
    \Bigg\}
    \\
    &\le
    \sup\left\{
    \frac{1}{\lambda n^{2}}
    \lVert \X (\vu-\valph)\rVert^{2}
    \,\middle|\,
    \vu,\valph\in[0,1]^{n}
    \right\}
    \\
    &=
    \sup\left\{
    \frac{1}{\lambda n^{2}}
    \lVert \X\vbeta\rVert^{2}
    \,\middle|\,
    \vbeta\in[-1,+1]^{n}
    \right\}
\end{tsaligned}
We now use the fact that $\forall \vbeta\in[-1,+1]^{n}$, 
\begin{tsaligned}
  \lVert\X\vbeta\rVert
  =
  \lVert\sum_{i=1}^{n}\beta_{i}\x_{i}\rVert
  \le
  \sum_{i=1}^{n}\lVert\beta_{i}\x_{i}\rVert  
  \le
  \sum_{i=1}^{n}\lVert\x_{i}\rVert
  \le
  n R, 
\end{tsaligned}
to get
\begin{tsaligned}
  C_{F}
  &\le
   \frac{1}{\lambda n^{2}}(nR)^{2} = \frac{R^{2}}{\lambda}. 
\end{tsaligned}
\qed

\section{Details on Experimental Results}
\subsection{Additional Experimental Results for Convergence}
\begin{figure*}[t!]
  \centering
  \begin{tabular}{lll}
    (a) $\lambda=10^{-6}/n$ &
    (b) $\lambda=10^{-4}/n$ &
    (c) $\lambda=10^{-2}/n$ 
    \\
    \includegraphics[width=0.24\linewidth]{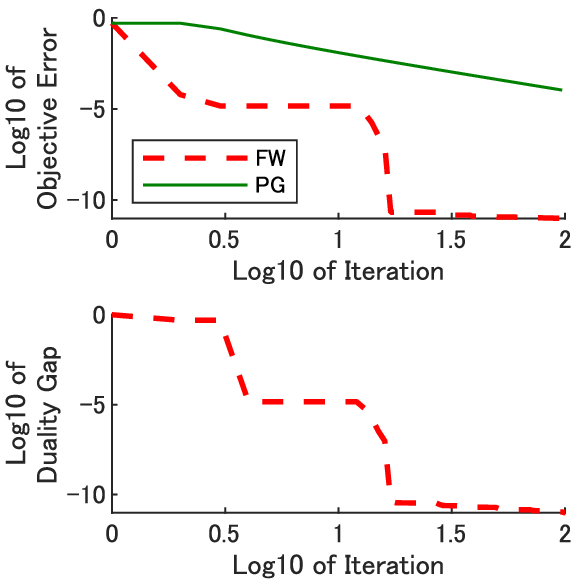}    
    &
    \includegraphics[width=0.24\linewidth]{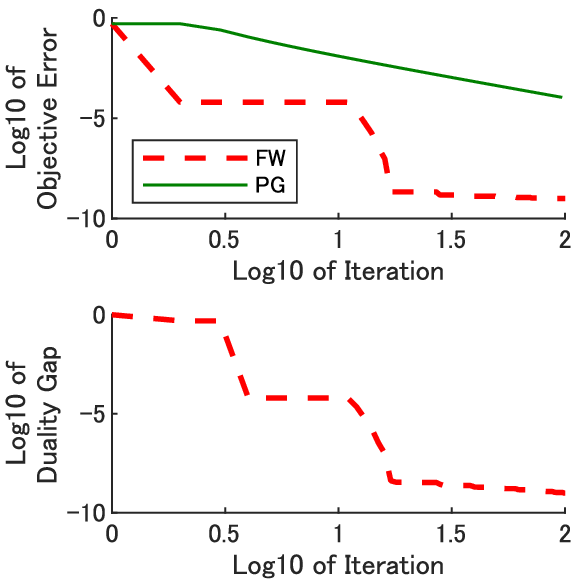}    
    &
    \includegraphics[width=0.24\linewidth]{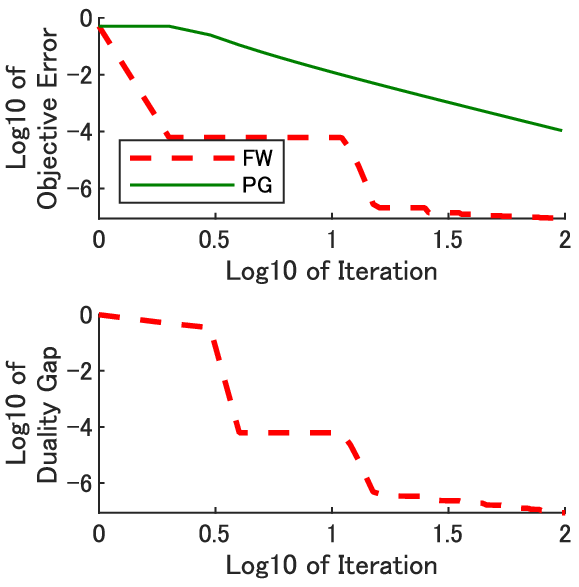}
    \\
    \\
  \end{tabular}
  \caption{
    Convergence on MNIST. 
    \label{fig:demo896-mnist}
    }
\end{figure*}

The convergence behaviors on MNIST are demonstrated in Figure~\ref{fig:demo896-mnist}. The experimental settings are described in the main text. 
\begin{figure*}[t!]
  \centering
  \begin{tabular}{lll}
    \\
    (a) $\lambda=10^{-6}/n$ &
    (b) $\lambda=10^{-4}/n$ &
    (c) $\lambda=10^{-2}/n$ 
    \\ \\
    \includegraphics[width=0.30\linewidth]{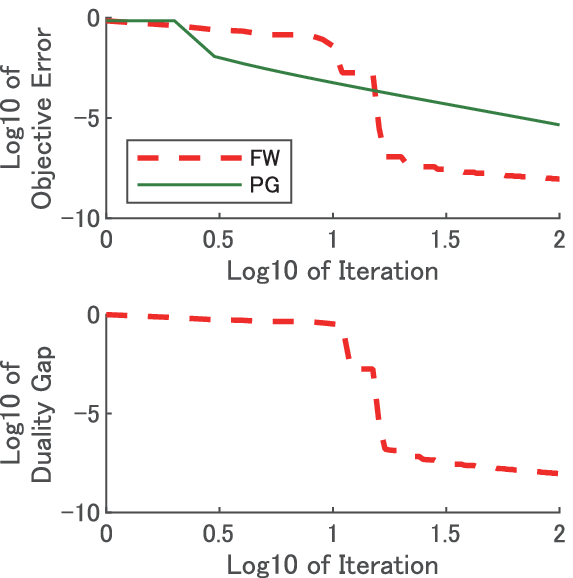}
    &
    \includegraphics[width=0.30\linewidth]{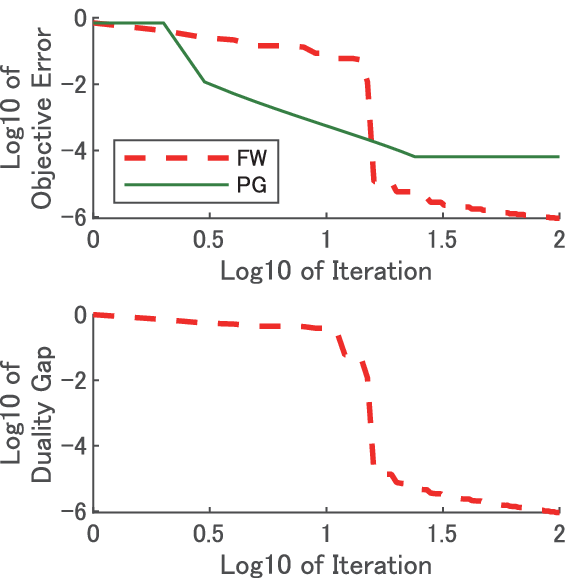}
    &
    \includegraphics[width=0.30\linewidth]{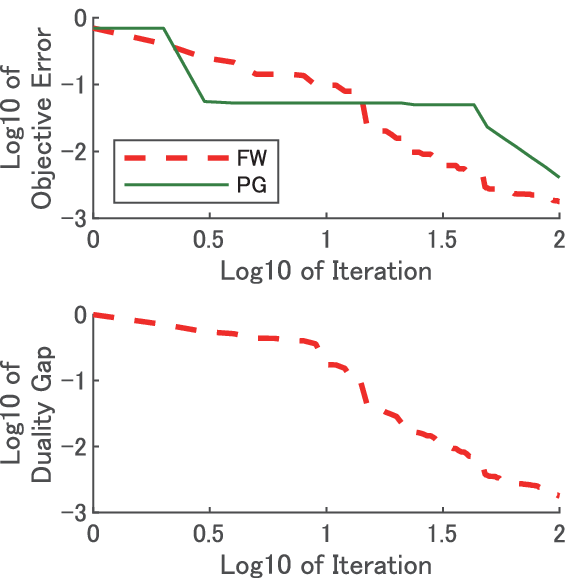}
    \\
    \\
  \end{tabular}
  \caption{
    Convergence on Cora.
    FW and PG are the abbreviations of Frank-Wolfe and projected gradient, respectively.  
    \label{fig:demo897-cora}
    }
\end{figure*}

\subsection{Additional Experimental Results for Amino Acid Sequence Classification}
\label{ss:protein}

\begin{figure}[t!]
  \centering
  \begin{tabular}{l}
    \includegraphics[width=0.29\linewidth]{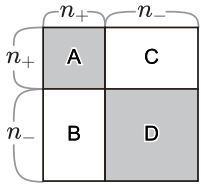}
  \end{tabular}
  \caption{
    Training data and sign constraints for SVM-pairwise.
In SVM-pairwise approach, features are sequence similarities to training data. Feature vectors for training SVM are columns of a square similarity matrix among training data. For simplicity, we assume that amino acid sequences are re-ordered so that first $n_{+}$ positive sequences are followed by $n_{-}$ negative sequences. In this assumption, the sub-matrix $\sfA$ of the similarity matrix is among positive examples, $\sfD$ is among negatives, $\sfB$ and $\sfC$ are of positives against negatives. Weight coefficients for sub-matrices $\sfA$ and $\sfC$ correspond to non-negative sign constraints~$\cI_{+}$, whereas weight coefficients for $\sfB$ and $\sfD$ correspond to non-positive sign constraints~$\cI_{-}$.     
    \label{fig:svmpairwise}
    }
\end{figure}

\begin{table*}[t!]
  \centering
  \caption{Details on ROC scores for amino acid sequence classification.
    \label{tab:demo388-07}}
  {
\begin{tabular}{|c|ccccc|}
  \hline & \multicolumn{5}{c|}{Conventional SVM-pairwise} \\
  & $\lambda=10^{-6}/n$ & $\lambda=10^{-4}/n$ & $\lambda=10^{-2}/n$ & $\lambda=10^{0}/n$ & $\lambda=10^{2}/n$ \\ 
  \hline
1 & 0.734 (0.010)  & 0.734 (0.010)  & 0.734 (0.010)  & 0.732 (0.010)  & 0.626 (0.011) \\
2 & 0.681 (0.022)  & 0.681 (0.022)  & 0.681 (0.022)  & 0.675 (0.020)  & 0.544 (0.023) \\
3 & 0.718 (0.011)  & 0.718 (0.011)  & 0.718 (0.011)  & 0.715 (0.010)  & 0.548 (0.014) \\
4 & 0.743 (0.009)  & 0.743 (0.009)  & 0.743 (0.009)  & 0.739 (0.008)  & 0.525 (0.009) \\
5 & 0.718 (0.011)  & 0.718 (0.011)  & 0.718 (0.011)  & 0.697 (0.010)  & 0.468 (0.021) \\
6 & 0.639 (0.012)  & 0.639 (0.012)  & 0.639 (0.012)  & 0.625 (0.012)  & 0.519 (0.011) \\
7 & 0.626 (0.029)  & 0.626 (0.029)  & 0.626 (0.029)  & 0.613 (0.026)  & 0.571 (0.015) \\
8 & 0.667 (0.022)  & 0.667 (0.022)  & 0.667 (0.022)  & 0.665 (0.022)  & 0.633 (0.028) \\
9 & 0.585 (0.024)  & 0.585 (0.024)  & 0.585 (0.024)  & 0.579 (0.020)  & 0.492 (0.024) \\
10 & 0.712 (0.013)  & 0.712 (0.013)  & 0.712 (0.013)  & 0.708 (0.012)  & 0.578 (0.011) \\
11 & 0.526 (0.025)  & 0.526 (0.025)  & 0.526 (0.025)  & 0.495 (0.024)  & 0.462 (0.016) \\
12 & 0.901 (0.013)  & 0.901 (0.013)  & 0.901 (0.013)  & 0.899 (0.013)  & 0.764 (0.015) \\
\hline
\hline & \multicolumn{5}{c|}{Sign Constrained SVM-pairwise} \\
  & $\lambda=10^{-6}/n$ & $\lambda=10^{-4}/n$ & $\lambda=10^{-2}/n$ & $\lambda=10^{0}/n$ & $\lambda=10^{2}/n$ \\
  \hline
1 & \underbar{0.752} (0.011)  & \underbar{0.752} (0.011)  & \underbar{0.752} (0.011)  & \underbar{0.754} (0.012)  & \textbf{\underbar{0.755}} (0.011) \\
2 & \textbf{\underbar{0.749}} (0.020)  & \textbf{\underbar{0.749}} (0.020)  & \textbf{\underbar{0.749}} (0.020)  & 0.731 (0.020)  & 0.676 (0.018) \\
3 & \underbar{0.744} (0.009)  & \underbar{0.744} (0.009)  & \underbar{0.744} (0.009)  & \textbf{\underbar{0.745}} (0.008)  & 0.646 (0.010) \\
4 & \textbf{\underbar{0.769}} (0.009)  & \textbf{\underbar{0.769}} (0.009)  & \textbf{\underbar{0.769}} (0.009)  & 0.766 (0.009)  & 0.684 (0.010) \\
5 & \textbf{\underbar{0.781}} (0.015)  & \textbf{\underbar{0.781}} (0.015)  & \textbf{\underbar{0.781}} (0.015)  & 0.768 (0.014)  & 0.680 (0.010) \\
6 & \textbf{\underbar{0.699}} (0.009)  & \textbf{\underbar{0.699}} (0.009)  & \textbf{\underbar{0.699}} (0.009)  & \underbar{0.697} (0.011)  & 0.634 (0.008) \\
7 & \underbar{0.710} (0.021)  & \underbar{0.710} (0.021)  & \underbar{0.710} (0.021)  & \textbf{\underbar{0.713}} (0.020)  & 0.695 (0.010) \\
8 & \textbf{\underbar{0.728}} (0.015)  & \textbf{\underbar{0.728}} (0.015)  & \textbf{\underbar{0.728}} (0.015)  & \underbar{0.722} (0.019)  & 0.689 (0.025) \\
9 & \textbf{\underbar{0.670}} (0.027)  & \textbf{\underbar{0.670}} (0.027)  & \textbf{\underbar{0.670}} (0.027)  & 0.653 (0.027)  & 0.608 (0.025) \\
10 & \underbar{0.741} (0.016)  & \underbar{0.741} (0.016)  & \underbar{0.741} (0.016)  & \textbf{\underbar{0.743}} (0.017)  & 0.645 (0.014) \\
11 & \textbf{\underbar{0.564}} (0.018)  & \textbf{\underbar{0.564}} (0.018)  & \textbf{\underbar{0.564}} (0.018)  & 0.533 (0.020)  & 0.487 (0.018) \\
12 & \textbf{\underbar{0.909}} (0.012)  & \textbf{\underbar{0.909}} (0.012)  & \textbf{\underbar{0.909}} (0.012)  & \underbar{0.908} (0.012)  & 0.857 (0.013) \\
\hline \end{tabular}
  }
\end{table*}

Here we report more details of the experimental results on amino acid sequence classification. As described in Subsection~\ref{ss:svmpairwise}, the procedure for dividing a set of 3,583 proteins into training and testing subsets was repeated 10 times to obtain ten divisions. For each division and each of 12 classification tasks, SVM was trained with the SVM-pairwise framework under sign constraints. Regularization parameter $\lambda$ was varied with $\lambda=10^{-6}/n, 10^{-4}/n, 10^{-2}/n, 10^{0}/n, 10^{2}/n$. The lower table in Table~\ref{tab:demo388-07} shows the average ROC scores and the standard deviations over ten repetitions. Similar procedure with the conventional SVM-pairwise was performed to obtain the upper table. The bold-faced figures in the table represent the best performance for the binary classification task. The underlined figures indicate that the performance is not significantly different from the best performance. The prediction performances in Table~\ref{tab:demo387-06} are the best ROC scores of the sign-constrained SVM-pairwise and the best ones of the conventional SVM-pairwise. In this application, weaker regularization tends to get a better prediction performance both for the two methods.

\end{document}